**ARTICLE**

# A Large Language Model Approach to Educational Survey Feedback Analysis

Michael J. Parker[1] · Caitlin Anderson[1] · Claire Stone[2] · YeaRim Oh[2]

## Abstract

This paper assesses the potential for the large language models (LLMs) GPT-4 and GPT-3.5 to aid in deriving insight from education feedback surveys. Exploration of LLM use cases in education has focused on teaching and learning, with less exploration of capabilities in education feedback analysis. Survey analysis in education involves goals such as finding gaps in curricula or evaluating teachers, often requiring time-consuming manual processing of textual responses. LLMs have the potential to provide a flexible means of achieving these goals without specialized machine learning models or fine-tuning. We demonstrate a versatile approach to such goals by treating them as sequences of natural language processing (NLP) tasks including classification (multi-label, multi-class, and binary), extraction, thematic analysis, and sentiment analysis, each performed by LLM. We apply these workflows to a real-world dataset of 2500 end-of-course survey comments from biomedical science courses, and evaluate a zero-shot approach (i.e., requiring no examples or labeled training data) across all tasks, reflecting education settings, where labeled data is often scarce. By applying effective prompting practices, we achieve human-level performance on multiple tasks with GPT-4, enabling workflows necessary to achieve typical goals. We also show the potential of inspecting LLMs' chain-of-thought (CoT) reasoning for providing insight that may foster confidence in practice. Moreover, this study features development of a versatile set of classification categories, suitable for various course types (online, hybrid, or in-person) and amenable to customization. Our results suggest that LLMs can be used to derive a range of insights from survey text.

Extended author information available on the last page of the article

## Introduction

Surveys in education have long been used for course evaluation and structured evaluation of teaching (SET) (Diaz et al., 2022; Dommeyer et al., 2004). The history of education surveys has seen examination of the pros and cons, with some noting the subjectivity of students' perspectives and others pointing to the difficulties in gathering responses unbiased by low response rates and the challenges with analyzing qualitative survey responses in an objective way (McGourty et al., 2002; Shah & Pabel, 2019; Spooren et al., 2013; Stowell et al., 2012; Wallace et al., 2019; Wongsurawat, 2011). However, evidence has supported the validity of surveys, including qualitative and mixed methods approaches, for guiding changes to teaching practice and course design (Ferren & Aylesworth, 2001; Johnson & Onwuegbuzie, 2004; Lattuca & Domagal-Goldman, 2007; Marks et al., 2017; McKeachie, 1997; Mentkowski, 1991; Spooren et al., 2013).

Common use cases for end-of-course evaluations (the type we focus on in this paper) include improving teaching and learning outcomes, measuring course quality, evaluating teachers, and informing decision-making (Diaz et al., 2022; Flodén, 2017; Marsh & Roche, 1993; Moss & Hendry, 2002; Schulz et al., 2014). Marsh and Roche (1993) found that university teachers receiving feedback and evaluations improved significantly more than a control group. Teachers can use feedback to target specific areas for improvement, with the potential to lead to structural course changes (Flodén, 2017; Schulz et al., 2014). Flodén (2017) observed positive effects of student feedback, tending to push teaching towards more interactive formats like seminars and group work rather than lectures. Education administrators have used feedback for quality assurance and to drive strategic decision-making (Ferren & Aylesworth, 2001; Marginson & Considine, 2000; Mazzarol et al., 2003). The scope is not limited to higher education. Student feedback surveys have played an increasing role in teacher development and evaluation in K-12 in recent years (Schulz et al., 2014). A large study found student surveys to be reliable and predictive of a teacher's ability to improve student achievement (Kane et al., 2013).

While quantitative feedback may have advantages for simplicity of analysis, the qualitative feedback gathered through student comments has additional value. In a higher education setting, Alhija and Fresko (2009) examined written student comments from end-of-course evaluations. They found that qualitative comments focused on the course, the instructor, and the context, capturing information not found in quantitative ratings. Comments covered unique aspects and provided more specific feedback on the strengths and weaknesses of a course. They concluded that qualitative feedback can provide a more comprehensive view of a course's teaching. Shah and Pabel (2019) used qualitative student feedback comments to compare the experiences of online and in-person students at their institution. Based on their insights, they conclude that universities *must* analyze qualitative student comments, not just quantitative ratings, to truly understand and enhance the student experience, especially given the growth in online education. The importance of qualitative feedback has been examined across online, in-person, and blended or hybrid format courses (Aldeman & Branoff, 2021; Alhija & Fresko, 2009; Onan, 2021a).

Despite the utility of qualitative student feedback, significant challenges remain in putting its usage into practice (Richardson, 2005). Shah and Pabel (2019) note the success in use of quantitative data but point to limited prior progress in the analysis and practical use of qualitative feedback. Approaches for analysis have included traditional, manual approaches such as thematic analysis that rely on manual effort of annotating and coding survey responses (Braun & Clarke, 2006; Riger & Sigurvinsdottir, 2016). Manually coding and analyzing large volumes of open-ended survey responses or student feedback comments is extremely time-consuming and labor-intensive and may not always provide actionable suggestions for improvement (Shaik et al., 2023; Mattimoe et al., 2021; Nanda et al., 2021; Shah & Pabel, 2019). Maintaining consistent coding across large educational datasets can be challenging when done manually, especially if multiple researchers are involved (Shaik et al., 2023; Mattimoe et al., 2021).

Employing crowdworkers, for example via Amazon's Mechanical Turk platform, reduces the cost and time of manual annotation of qualitative data but does not solve some of the other issues. The quality of results may vary, particularly in cases where some degree of domain expertise is needed (Gilardi et al., 2023; Rother et al., 2021). Qualitative annotation tasks can be inherently subjective, leading to disagreements among crowdworkers and low inter-annotator agreement (Pradhan et al., 2022; Rother et al., 2021). In addition, a recent study (Veselovsky et al., 2023) provided evidence that a substantial fraction of crowdworkers used generative AI (Large Language Models, or LLMs) to assist with a summarization task, leading to a mix of results from humans and LLMs and raising doubt that crowdworkers will continue to be a reliable source of human annotations.

More recently, automated methods for analysis of qualitative data have relied on a variety of machine learning models, (Deepa et al., 2019; Onan, 2021b; Smith & Humphreys, 2006; Zhang et al., 2020). Machine learning approaches (discussed further below)—including unsupervised semantic mapping, topic modeling, and using of different forms of neural networks—have been applied to a range of tasks like clustering, summarization, entity extraction, and sentiment analysis (Gottipati et al., 2018; Hamzah et al., 2020; Nanda et al., 2021; Patil et al., 2019; Shaik et al., 2023). These approaches have shown promise in aiding analysis, but often require conditions that make their use less feasible to most educators, such as the need for significant technical resources, fine-tuning of models on volumes of pre-existing labeled data, use of separate models for the natural language processing tasks involved (impeding broader analyses), or the need for use of specialized software (Fan et al., 2015; Gottipati et al., 2018; Orescanin et al., 2023; Pyasi et al., 2018; Smith & Humphreys, 2006). These models have therefore generally been the domain of research, with a gap in widely accessible, practical methods for qualitative analysis of end-of-course surveys.

Large language models with generative AI capabilities have become more widely available, capable, and accessible (easier-to-use) in the last one to two years, but are underexplored in terms of use in survey analysis versus more specialized machine learning models. LLMs have the potential to circumvent many of the problems associated with specialized machine learning approaches and potentially democratize access to high quality qualitative survey analysis. For example, the ability to use such models through natural language instructions, via simple web interfaces, or at scale through application programming interfaces (APIs), lessens the need for dedicated machines or expensive software. However, a thorough analysis of the feasibility and evaluation of the quality of LLMs' results has not been performed to establish reliability and rigor in common qualitative survey analysis tasks. Our *main research question* was: *are large language models are at a stage where they can be effectively used across the broad range of tasks that are part of survey analysis?* To answer this main question, we propose the following related research questions:

- **Research question 1 (RQ1):** Can LLMs be used to perform multiple unstructured text analysis tasks on educational survey responses, including multi-label classification, multi-class classification, binary classification, extraction, inductive thematic analysis, and sentiment analysis?
- **Research question 2 (RQ2):** Can LLMs' chain-of-thought (a demonstration of the intermediate steps of how they arrive at their answers) be captured to provide a degree of transparency that may help foster confidence in real world usage? Can we demonstrate examples that show the potential for this use case?
- **Research question 3 (RQ3):** Is a zero-shot approach (not needing to provide hand-labeled examples) across all tasks, a scenario that mimics many real-world practical use cases in the education setting, capable of achieving performance comparable to human annotation?

As part of the evaluation process, we also developed a set of classification categories that can be applied to a variety of course types (online, hybrid, or in-person), and which are amenable to customization depending on specific requirements.

## Background

### Types of Tasks Associated with Analysis of Unstructured Survey Data

Analyzing survey textual responses to explore the high-level goals of educational stakeholders requires chaining together natural language processing (NLP) tasks in the form of workflows. Such workflows can be implemented with NLP tasks, including classification, extraction, and sentiment analysis, that form composable building blocks for similar workflows.

Classification of comments may be single-label (binary or multi-class, the latter involving classifying into one of a set of tags) or multi-label (classification of each comment with one or more of a set of tags). The tags (also called labels, classes, or categories) are frequently custom-chosen, reflecting the goals of a particular analysis (Goštautaitė & Sakalauskas, 2022). Often those doing the analysis have a specific objective or goal focus that they are investigating (e.g., suggestions for improvement), and text extraction is a useful technique for this purpose. Sentiment analysis can be used to lend nuance and insight to the quantitative ratings that are gathered through Likert scales or "star" ratings (Gottipati et al., 2017; Nitin et al., 2015).

A high-level breakdown of objectives and NLP tasks is shown in Table 1.

Table 1 NLP tasks that may be used for analysis of textual survey responses

| Objective | Question | NLP Tasks | Notes |
| --- | --- | --- | --- |
| High-level initial analysis | What did students say (and how did they feel about the course)? | Multi-label classification, inductive thematic analysis, sentiment analysis | Depends on whether analysis is top-down (using pre-determined labels or areas of interest) or bottom-up (deriving themes from scratch based on student comments) |
| Answering a focused question | What did students say about x (particular focus)? | Extraction | Results are amenable to multi-class classification or inductive thematic analysis |
| Quantification of textual survey responses | How many comments were there on each aspect? | Classification (binary, multi-label, or multi-class) | Helps the person performing analysis find themes of greater importance |

For discussion on the background of automated means of qualitative survey analysis, it is helpful to divide methods into those developed prior to the broad availability of the most recent generative AI versus those that make use of the recent generative AI advances in the form of LLMs. It is a potentially useful simplification to think of the pre-generative AI approaches as "language in, numbers out" and the later generative AI approaches as "language in, language out" in terms of how one interacts with the models. These approaches are distinguished below.

**Previous Machine Learning Approaches and Challenges in Analyzing Education Feedback**

For feature extraction from text, techniques like TF-IDF, and Word2Vec have been applied for short text classification and sentiment analysis (Deepa et al., 2019; Devlin et al., 2018; Onan, 2021a; Zhang et al., 2020). Topic modeling using latent semantic analysis or latent Dirichlet allocation has been useful for discovering themes and trends in collections of student feedback (Cunningham-Nelson et al., 2019; Perez-Encinas & Rodriguez-Pomeda, 2018; Unankard & Nadee, 2020). For evaluating text, sentiment analysis techniques like CNN and Bi-LSTM models have been used to classify student evaluations (Sindhu et al., 2019; Sutoyo et al., 2021). Overall, these techniques have shown utility for gaining insights from student feedback.

With the advent of recent machine learning (ML) techniques, great strides have been made in dealing with unstructured text. BERT (Bidirectional Encoder Representations from Transformers, Devlin et al., 2018) and related models allow for transformation of text passages into numerical formats (high dimensional dense vectors called embeddings) that are then amenable to classification via conventional ML methods such as logistic regression. Good results have been achieved in certain contexts using such models (Meidinger & Aßenmacher, 2021). Despite such advances, challenges remain that present obstacles to routine use of such models in practice.

Specialized ML models often require a "fine-tuning" process using labeled data (data that human annotators have classified) to best adapt to a specific use case. Depending on the amount of human labeling needed, this aspect may provide a stumbling block based on the time and effort involved. Although there are many examples of labeled datasets (Papers With Code datasets, n.d.; Hugging Face datasets, n.d.; Kastrati et al., 2020a, b), real-world use cases often rely on custom labels for which there is no pre-existing labeled data for fine-tuning. Even supposing such fine-tuning takes place, there are additional barriers to practical use of this technology.

One such barrier is that multiple distinct AI models may be needed, depending on the range of tasks. The model that is suitable for classification may not be the same one that performs text extraction, and each model may need its own fine-tuning or adaptation.

Even for a core task like classification, there are a number of challenges. Difficulty of classification increases in situations where multiple labels may concurrently be assigned to the same survey comment, often leading to a degree of inter-rater disagreement even among highly-skilled human annotators who have high familiarity with the domain. Other challenges include data imbalance, multi-topic comments, and domain-specific terminology (Edalati et al., 2022; Shaik et al., 2022).

In classifying unstructured textual feedback, data imbalance exists when the labels chosen are not attributable in equal proportions across a dataset; some labels may be comparatively rare. If there are few examples of particular labels, this scarcity can create difficulties in training machine learning models that classify new comments. If human labeling is being used as ground truth, rarity of certain labels may require labeling a larger set of feedback to enable training an ML classifier. Techniques addressing data imbalance include synthesizing new training examples for the minority class through data augmentation or by oversampling the minority class instances through techniques like SMOTE (Kennedy et al., 2024). Other methods involve modifying the learning algorithms to assign higher misclassification costs to minority class examples, such that the model parameters are affected more by rare class examples, and ensemble methods trained on resampled versions of the data (Johnson & Khoshgoftaar, 2019; Shah & Ali, 2023).

Another challenge is that of multi-topic comments. Depending on how feedback is collected and how open-ended the survey questions are, students may provide feedback that encompasses multiple topics (for example, "I found the quizzes incredibly difficult, but the teacher was great and I felt I got what I paid for. If I had had more time to complete the course, this would have been even better."). Such multi-topic comments present a challenge for ML techniques based on embeddings (dense vector representations) derived from models such as BERT (or BERT related, such as Sentence-BERT, Reimers & Gurevych, 2019), given that the embedding of a comment is related to the comment's semantic meaning. A comment with multiple topics may have an embedding that doesn't adequately localize to the semantic "neighborhood" of any of the individual topics associated with that comment, decreasing the performance of downstream classifiers.

Use of context-specific, specialized terms in the text data, known as domain-specific language, can also decrease the performance of ML techniques. Deep learning models like BERT that perform feature selection by creating embeddings have been pre-trained on a large corpus of text, usually publicly accessible and mostly from the internet. Depending on the pre-training, terms specific to a specialized domain such as immunology or biomedical engineering may not have been seen during training, or seen only in very limited quantities. In those cases, the pre-trained model cannot adequately capture the semantics of such terms via its embeddings, again impacting the performance of downstream applications such as classification and clustering that may rely on those embeddings. For example, Lee et al. (2020) discuss how the original BERT was pre-trained on general domain corpora like Wikipedia and BookCorpus, which lack sufficient technical biomedical vocabulary and writing styles. This leads to poor performance on biomedical NLP tasks. Gu et al. (2021) analyze how continual pre-training of BERT on biomedical corpora like PubMed abstracts and PMC full-text articles significantly improves performance on downstream biomedical tasks compared to the general BERT model. The key reasons cited for why BERT models may struggle with domain-specific language are the lack of domain-specific vocabulary, writing conventions, and linguistic patterns in the original BERT pre-training corpus, which leads to poor representations for technical terminology and jargon when applied to domain tasks without additional in-domain pre-training.

In sentiment analysis, pre-trained sentiment analysis models may not adapt well to settings where it is important to take into account the context. For example, in analyzing comments from biomedical science courses that cover cancer as a topic, learners' comments may include the words 'cancer' or 'oncology' or 'tumor', simply as referring to parts of the curriculum. These comments may end up being classified as negative even by a state-of-the-art existing model, given that discussions of cancers and tumors in many training datasets (often from internet settings) may be in the context of negative emotions being expressed.

Finally, a common challenge is that of lack of interpretability of results coming from specialized machine learning models (Hassija et al., 2024). Although there has been significant work on approaches like visualizing factors that contribute to a neural network-based model's predictions, complex models may still be viewed as "black boxes" by downstream users in areas like education, with this perception potentially inhibiting usage.

**LLM Background and Related Research**

Education feedback analysis seeks to extract insights from open-ended written responses, such as student surveys or teacher evaluations, and automated techniques can be seen as a particular application of the broader field of natural language processing (Shaik et al., 2022). The introduction of transformer-based neural network architectures in 2017 led to an explosion of new AI models for NLP with increasing capabilities. BERT (mentioned above) was developed shortly thereafter (2018), with multiple related models (e.g., RoBERTa) being further developed over the last five years, with effectiveness at various NLP tasks that often exceeded those of pre-transformer models. Such models have been applied to a wide range of tasks, both with fine-tuning and without.

Large language models are neural networks based on transformer architectures, including not only those in the BERT lineage but also other models such as GPT-2, GPT-3, T5, and many others, with tremendous scale in terms of the number of model parameters (billions and sometimes trillions) and the internet scale volume of text on which they are trained (billions or even trillions of tokens, with tokens being numerical representations of words or parts of words). BERT (the large variant) has approximately 345 million parameters and was trained on about 3.3 billion words; in comparison, GPT-3 has 175 billion parameters and was trained on approximately 500 billion tokens (approximately 375 billion words). Models like GPT-3.5 and GPT-4 are proprietary, and the number of parameters and the amount of training data are unknown, although there are estimates that GPT-4 uses approximately 1.8 trillion parameters and was trained on approximately 13 trillion tokens, with GPT-3.5 somewhere in between GPT-3 and GPT-4 (Schreiner, 2023).

It is important as well to distinguish between BERT, along with related models like RoBERTa and SentenceTransformers, and generative models like GPT-3.5 and GPT-4. While all of these are transformer models (a type of neural network architecture),

BERT is known as an "encoder-only" model, more suitable for feature extraction, clustering, and a variety of use cases making use of the resulting embeddings (numerical representations of language, discussed above). The GPT models and other recent generative models are "decoder-only, auto-regressive" models, with different strengths, including the ability to generate complex text. These models are also trained on an order of magnitude more text, which lends to their capabilities as well.

These generative AI models have the capability to do tasks like summarization, translation, and generation of high-quality text output. As their scale has grown, the range of tasks of which they have shown to be capable has increased, along with a level of performance that has surprised many. With the recent popularization and wider spread availability of LLMs, in part due to ChatGPT, with its underlying GPT-3.5 and GPT-4 models, as well as other LLMs like Claude (Anthropic), Command (Cohere), Bard/Gemini (Google), Llama (Meta), and a range of open-source models, interest has grown in applying these to use cases like analysis of short text comments such as are seen in Tweets (Törnberg, 2023), customer feedback, and survey feedback (Jansen et al., 2023; Masala et al., 2021).

Multiple recent studies have examined using ChatGPT for text annotation and classification tasks, with mixed results based on variations in prompts, datasets, parameters, and complexity of tasks. Reiss (2023) focused on sensitivity to the prompts and parameters used in classification, in the context of performing classification on a German dataset. Pangakis et al. (2023) argues that researchers using LLMs for annotation must validate against human annotation to show that LLMs are effective for particular tasks and types of datasets, given that there is variation in the quality of prompts, the complexity of the data, and the difficulty of the tasks. Other studies (Gilardi et al., 2023; Huang et al., 2023) demonstrate the potential for ChatGPT to perform text annotation or provide natural language explanations at levels approaching or matching those of humans.

Despite explorations like those mentioned above, research to date has not focused on the feasibility and quality of LLMs' results in performing a broad array of common qualitative education survey analysis tasks, leaving a gap that we focus on in this study. For example, a review published in 2024 focusing on natural language processing of students' feedback to instructors makes no mention of studies using LLMs for this purpose (Sunar & Khalid, 2024). Prior work has primarily focused on the use of encoder models like BERT for their clustering and feature extraction capabilities and have not explored the current generation of decoder-only auto-regressive models like the GPT models mentioned above. Our contribution is to explore and evaluate the capabilities of the most recent generation of LLMs for educational survey feedback analysis.

## Methods

### Survey Data Used for Evaluation

2,500 survey responses (625 for each of four open-ended questions) were selected at random from a larger set of survey responses (50,525 total responses as of April 16, 2023, when data collection was performed, with 14,359 responses for Q1 (see below), 13,654 responses for Q2 (see below), 12,825 responses for Q3 (see below), and 9,687 responses for Q4 (see below)) received as end-of-course feedback on a range of online biomedical science courses. The courses included 365 course iterations total across 15 different courses (e.g. Pharmacology, Genetics, Immunology, etc.). An additional 2,000 survey comments were chosen as a development set that could be used for LLM prompt tuning. The courses all use a single, uniform end-of-course survey. In addition to quantitative ratings (e.g., net promoter scores) and optional demographic data, the survey included open-ended text responses to four questions/directives:

– "Please describe the best parts of this course." [Q1]
– "What parts of the experience enhanced your learning of the concepts most?" [Q2]
– "What can we do to improve this course?" [Q3]
– "Please provide any further suggestions, comments, or ideas you have for this series." [Q4]

On average, learners answered approximately two of the four questions. The shortest responses containing content were one word, and the longest responses were several paragraphs. Example survey responses are shown in Table 2.

Survey responses were collected via Qualtrics, with minor processing with Pandas 2.0.1 for elimination of leading and trailing white space and automated removal of responses with no content (NA or None variants).

Survey responses were inspected manually and via named entity recognition (NER), running locally, to ensure that no private or sensitive information was transmitted to publicly available LLMs.

**Development of Course Tagging System**

We spent considerable time developing and testing a set of labels that would work well not only for online courses like those that the survey responses in this paper were a part of, but also other types of educational offerings. The motivation for the choice of labels was based on the functional areas of the team creating and delivering the courses. This team had separate functions (sub-teams) for curriculum/teaching, logistics and operations (e.g. enrollment and technical aspects of course delivery), creative media/visual asset development (art and video creation), technology (learning management system and platform development and maintenance), administration/leadership (analysis of feedback for improvement and future feature/course requests), and support (solving learner issues and handling inquiries). While most course teams at universities or online learning providers may not have all these functions separately, they are generally present in some fashion even if a single individual (for example, an instructor) covers multiple functions. Tags were therefore designed to cover functional areas and enable identification of feedback of interest for each team's quality improvement and planning processes.

**Table 2** Example actual survey responses

| [Q1] responses | [Q2] responses | [Q3] responses | [Q4] responses |
| --- | --- | --- | --- |
| "The teachers they are incredible and their fascination about this topic make it more interesting." | "the whole concept, the short videos with the explanations written down and then the interactive modules" | "Implement more checkpoints that review previous material throughout the course." | "I really enjoyed the course and learned a lot of applicable information for my job. I would have like a little more time between new releases of information. It would also be nice to have a live question/answer session." |
| "the structure of this course is just great. however i would love to have the chance to repeat all the modules as i am from a very different background." | "The quizzes after each module made me think about the material I just learned." | "The course was fantastic and informative. However, I had to rewatch the videos several times to write down everything that is said. I learn best by looking at the words. The videos should come with either a transcript or written words or some sort that convey the same information" | "A visit to meet the tutors and a summary discussion on location would be fabulous—I am aware not many people would make it, but a thought nonetheless." |

The label development process started with a much larger set of labels (71 total). Given that each survey response could cover multiple topics, the task was to assign as many labels to each survey response as were applicable (a multi-label classification task). The four authors (all of whom have been involved in either course development or delivery for multiple years and can be considered domain experts in the course resources and processes) each labeled a test set of 2,000 survey responses (from the same educational program overall, but distinct from the set of 2,500 comments ultimately labeled), with resulting relatively low inter-rater agreement. Based on this experiment, tag categories were combined to arrive at a much smaller set of generalizable tags (see Table 3). These were still applied in a multi-label classification approach, with each survey response potentially receiving multiple labels. For example, the example comment mentioned earlier ("I found the quizzes incredibly difficult, but the teacher was great and I felt I got what I paid for. If I had had more time to complete the course, this would have been even better.") might simultaneously receive labels of "assessment", "teaching", and "course logistics and fit". In addition, best practices were followed to ensure generalizability (University of Wisconsin—Madison, n.d.; UC Berkeley Center for Teaching & Learning, n.d.; Brennan & Williams, 2004; Medina et al., 2019).

A one to three sentence description of each tag was created to provide guidance so that tags could be applied appropriately in testing rounds. The intent is also that others can adapt these same tags by modifying the description portion for their own purposes. The same descriptions that served as context for the human annotators were also used in the prompts for the LLMs in the multi-label classification task as a form of deductive thematic analysis.

Table 3 Final tags and descriptions

| Tag | Description |
| --- | --- |
| Course logistics and fit | course delivery (policy, support), cost, difficulty, time commitment, grading, credit, schedule, user fit, access, background (e.g., prereqs and appropriateness of course level) |
| Curriculum | course content, curriculum, specific topics, course structure. This focuses on the content and the pedagogical structure of the content, including flow and organization. This also includes applied material such as clinical cases and case studies. Includes references to pre-recorded discussions between experts or between a doctor and a patient. Includes specific suggestions for additional courses or content |
| Teaching modality | video, visual, interactive, animation, step-by-step, deep dive, background builder (the format rather than the content/topic) |
| Teaching | instructors, quality of teaching and explanations |
| Assessment | quizzes, exams |
| Resources | note taking tools, study guides, notepads, readings. Includes other potential static resources like downloadable video transcripts |
| Peer and teacher interaction | includes chances for the student to interact with another person in the course (teacher or student). This includes discussion forums, teacher-student or student–student interactions. Includes requests for live sessions with teachers or live office hours |
| Other | catch-all for the rarer aspects that we'll encounter and also the 'na', 'thank you', etc. comments that don't really belong in the above bins. Also for sufficiently general comments like 'all the course was terrific' that can't be narrowed down to one of the other categories |

We then iteratively tested the new, much smaller set of tags on several sets of 100 survey responses, with all four authors independently tagging the same entries, followed by examination of inter-rater agreement. This yielded good results. With this set of tags, we then independently labeled 2,500 survey responses, and evaluated inter-rater agreement using Jaccard similarity coefficient between pairs of raters and averaged across all pairs of raters. Jaccard similarity was chosen because it is one of the most common metrics for multi-label classification. Other common multi-label classification metrics (e.g., Hamming distance) were also calculated and are provided in the appendix (see discussion of metrics below).

**LLM Processing**

All LLM tasks were performed via calls to the OpenAI API endpoints. GPT-3.5 (model: gpt-3.5-turbo-0301) and GPT-4 (model: gpt-4–0314) were used for the multi-label classification task; all other tasks described used GPT-3.5 (model: gpt-3.5-turbo-0613) and GPT-4 (model: gpt-4–0613). All tests were run with calls to the models' asynchronous endpoints and used a temperature of 0 with other parameters set to their default values, other than the *functions* parameter and the *function_call* parameter, which were set to specify the applicable function schema and the function name where applicable. "Function calling", a capability specific to these models, was used to generate the JSON structured output for all tasks. Prompts used (see Electronic Supplement) involved function schemas, as well as the system and user messages to the model.

For the LLM approach to the multi-label classification task, the multi-class classification task for extracted excerpts, the binary classification task, and the sentiment analysis task, zero-shot chain-of-thought (CoT) prompting was used (where a model is prompted to reason step-by-step but without examples of such reasoning provided) (Kojima et al., 2022; Wei et al., 2022). In addition to use of CoT enhancing the accuracy of the model output, the reasoning was included in the output to allow for error analysis and prompt tuning, as well as to allow inspection of the model's reasoning, something potentially helpful for those using the results in practice. For sentiment analysis, we had the LLM output a sentiment classification based on the possible categories 'negative', 'slightly negative', 'neutral', 'slightly positive', and 'positive', along with its reasoning.

For the LLM approach to inductive thematic analysis of survey responses, a two-step approach was used. The first step involved prompting the LLM to derive themes representing feedback from multiple students and summarize the themes. This step was run in parallel on batches of survey responses that would fit within the model's context window. The second step involved prompting the LLM to coalesce derived themes based on similarity to arrive at a final set of themes and descriptions. These steps could be considered analogous to part of the human inductive thematic analysis qualitative analysis workflow (Braun & Clarke, 2006).

Various prompting techniques were used to improve the results. These included:

1) Zero-shot CoT—This technique involves asking the model to think step-by-step to arrive at a correct result and to provide its detailed reasoning. In the absence of providing examples of CoT reasoning in the prompt, this type of prompting is categorized as zero-shot.
2) Prompt tuning via inspection of CoT reasoning—In testing, error analysis was supplemented with inspection of CoT reasoning to help discern where prompts might need refinement. As prompts were updated, we observed corresponding changes in the output and the stated reasoning, with improvement in the development set metrics.
3) Additional descriptive context for labels—Given that there was no fine-tuning to allow the model to learn the appropriate context and meaning of labels, we added context to prompts in the form of definitions for each label and the types of elements for which each label applied.
4) Additional context through injection of the survey questions into the prompt—Inclusion of additional context, such as the survey question that a given comment is in reply to, may improve the performance of LLMs and was used in this study.
5) Use of function calling for reliable structured output—This technique is specific to the GPT-3.5 and GPT-4 models, for which models from the June 2023 checkpoint (0613) onward have been fine-tuned to enable structured output (e.g., JSON) when provided with information about a function schema that could be called with the output.
6) Memetic proxy, also known as the persona pattern (Reynolds & McDonell, 2021; White et al., 2023)—Asking the LLM to act as a certain persona, for example as an expert in survey analysis tasks, has been described as another way to improve results, potentially by helping the model access a portion of its memory that holds higher quality examples of the task at hand.

**Other Models**

In addition to comparison to human ground truth labels, for multi-label classification, comparison was made to SetFit (Tunstall et al., 2022), a SentenceTransformers finetuning approach based on Sentence-BERT and requiring very little labeled data; for sentiment analysis, comparison was made to a publicly available RoBERTa-based model trained on 124 M Tweets. Both are encoder-only models (distinct from generative AI models like GPT-4 and GPT-3.5), with the embeddings used in fine-tuning to enable classification. These comparisons provide some context for the LLMs' performance relative to recent specialized models.

**Evaluation Metrics**

Scikit-learn 1.2.0 was used for statistical tests, along with numpy 1.23.5 and Pandas 2.0.1 for data analysis. Weights & Biases was used for tracking of model evaluation results.

For the multi-label classification task, model results were compared to the human ground truth labels. Two ways were used to arrive at ground truth labels aggregating results from multiple annotators: 1) using **consensus rows:** only the subset of survey responses (dataset rows) where all four annotators had majority agreement on all selected tags were kept; and 2) using **consensus labels:** all survey responses were kept but only labels with majority agreement were chosen as selected.

To fine-tune the SetFit model, we used a portion of each ground truth dataset (the first 20 examples for each label). Those examples were omitted from the test set, leaving 2,359 rows in the consensus labels test set and 1,489 rows in the consensus rows test set.

For each of the above scenarios, model results for multi-label classification were evaluated against aggregated human annotator results via the following metrics: 1) Jaccard similarity coefficient, comparing the model against aggregated human results for each row (survey response) and then averaged over all rows; 2) average precision per tag; 3) average recall per tag; 4) macro average precision, recall, and F1 score across all tags; 5) micro average precision, recall, and F1 score across all tags; 6) Hamming loss; and 7) subset accuracy.

For the binary classification task, accuracy, precision, recall, and F1 score were calculated, comparing the model results to one expert human annotator.

For the extraction task, extracted excerpts were evaluated by GPT-4 using a rubric created specifically for this task, examining performance on multiple aspects of performance, including the presence of excerpts that were not exact quotes from the original (part of the original extraction instructions), the completeness of capturing relevant excerpts, the presence of excerpts irrelevant to the initial goal focus, the inclusion of relevant context from the original comment, and several others. The results were also evaluated by human annotation to determine the presence of hallucinations (excerpts that were substantial changes from the original survey responses, rather than just changes in punctuation, spelling, or capitalization), with the percent of the total number of excerpts representing hallucinations being reported.

For the inductive thematic analysis task, there is not an accepted evaluation method given that this is a complex, compound task, and evaluation consisted of inspecting the derived themes and descriptions as well as inspecting the results of the associated multi-label classification step.

The sentiment analysis results of GPT-3.5 and GPT-4 were compared to those of a RoBERTa sentiment classifier trained on ~ 124 million tweets (Hugging Face cardiffnlp/twitter-roberta-base-sentiment-latest, 2022; Loureiro et al., 2022), as well as to results from a human annotator, with accuracy, precision, recall, and F1 scores reported for the prediction of sentiment as negative, neutral, or positive. Comparison was made by grouping 'negative' and 'slightly negative' into a single class, keeping 'neutral' as its own class, and grouping 'positive' and 'slightly positive' into a single class to allow for comparison across sentiment analysis methods. The RoBERTa classifier produced a dictionary with negative, neutral, and positive classes, with probabilities summing to 1.0. The class with the maximum probability score was chosen as the label for comparison to the human annotations.

# Results

In this section, we organize the results in relation to the three research questions. The first research question pertains to whether LLMs can perform multiple text analysis tasks on unstructured survey responses, including multi-label classification, multi-class classification, binary classification, extraction, inductive thematic analysis, and sentiment analysis. The section below on *approach to LLM workflows* answers this research question. The second research question explores whether LLMs' chain-of-thought (a demonstration of the intermediate steps of how they arrive at their answers) can be captured to provide a degree of transparency that may help foster confidence in real world usage. In the section below on *chain-of-thought reasoning*, we demonstrate examples that show the potential for this use case. The final research question asks whether a zero-shot approach (not needing to provide hand-labeled examples) across all tasks can achieve performance comparable to human annotation. The section below on *individual NLP task evaluations* answers this research question.

For the examples, we use GPT-4 as the LLM; the evaluations compare GPT-4 and GPT-3.5 as well as the other models used.

## Approach to LLM Workflows

The main types of workflows demonstrated support the goals shown in Table 1 of 1) high-level analysis, in which the desire is to understand the main categories and areas of emphasis across all student feedback, or 2) more focused analysis, e.g., answering specific questions about a particular aspect of a course. In both cases, quantification of results is a consideration, which is supported by classification tasks.

For initial, high-level analysis across the entire set of survey comments, we demonstrate two approaches: 1) inductive thematic analysis, a "bottom-up" approach supporting the use case where no predetermined labels (areas of interest) have been defined, similar to topic modeling, and 2) multi-label classification using predefined labels, a "top-down" approach, also referred to as deductive thematic analysis. When categories of interest are known in advance, multi-label classification is an appropriate first step, binning survey responses into relevant categories that provide a sense of the type of feedback learners are providing. These categories also provide groupings of comments for further focused analysis (e.g., via extraction), as well as allow for quantification based on the number of comments labeled with each category.

For focused analysis, in which there is a specific question or goal for the analysis, not necessarily known in advance, we demonstrate extraction as a key step, followed by either a classification step or thematic analysis. To provide output for further downstream analysis and quantification, multi-class classification can be used as a step, as demonstrated here with the generalizable set of labels used in this study, or with an adapted or fully customized version for one's own use case. This step is shown used after extraction, given that short excerpts are more likely to be adequately classified with a single label versus multi-sentence comments. The output of other forms of classification (binary or multi-label) also lends itself well to quantification of results.

Sentiment analysis was applied as a final step for workflows where finding positive or negative excerpts was of interest, as demonstrated in the example related to the level of difficulty of the course.

Although the full model responses were in JSON format, only the relevant output text is shown for brevity and clarity.

**Workflow Examples**

**Example 1—High-Level Analysis by Inductive Thematic Analysis ("Bottom-Up" Approach)**

A workflow for finding and summarizing the main themes (ideas expressed by multiple students) of survey responses is shown in Fig. 1, and consists of three LLM steps: 1) themes are first derived and summarized for batches of comments, each of which is sized to fit within the context window of the model used; 2) comments are classified using the derived themes; and 3) sets of themes from these batches are coalesced to arrive at a final set of themes. Additionally, label counts are aggregated from the themes that were combined. In qualitative research, steps 1 and 3 are called inductive thematic analysis; this is similar to topic modeling, in that themes are inductively derived from comments. In general, depending on the input size (context window) for the model used (8 K tokens in this example) and the number of comments being analyzed, dividing into batches and coalescing the themes from each batch may be unnecessary.

Results for running this process on the 625 comments from Q1 ('Please describe the best parts of this course') are shown in Fig. 1. The number of comments that the LLM identified as corresponding to each theme is shown, along with the theme titles and descriptions.

**Example 2—High-Level Analysis by Categorizing Student Comments ("Top-Down" Approach)**

Multi-label classification of survey responses, using the set of predetermined labels developed for this study (Table 3) was run on the 625 comments from Q1 ('Please describe the best parts of this course') and results are shown in Fig. 2. The categorized comments can be used for analysis (for example, comparing the categorization of responses to 'Please describe the best parts of this course' to the categorization of responses to 'What can we do to improve this course?') or as a starting point for further downstream tasks.

**Example 3—Finding Suggestions for Improvement**

A workflow for finding and quantifying suggestions for course improvement is shown in Fig. 3, and consists of extraction of relevant excerpts, followed by multi-class classification, based on the labels in Table 3, to facilitate quantification as well as routing of comments to the appropriate stakeholders. Excerpts resulting from the extraction step were assumed to be focused enough that they could each be categorized with a single class from among the pre-existing labels in Table 3.

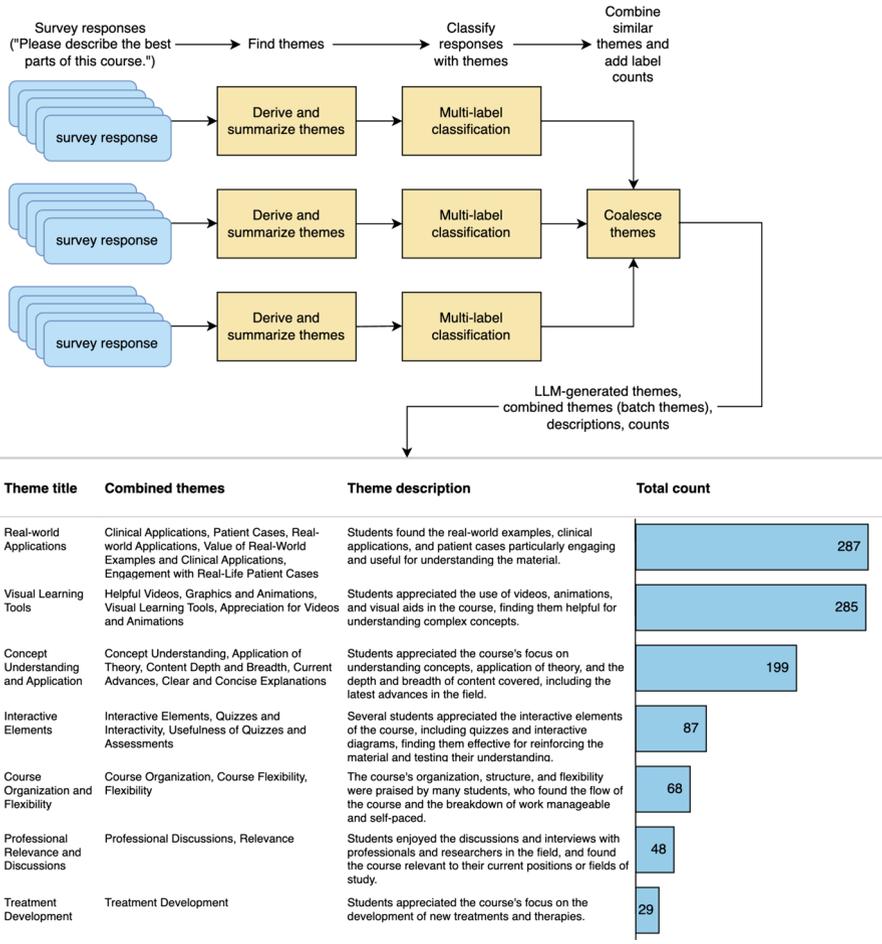

**Fig. 1** Derivation of themes from student comments (results shown using GPT-4)

Results for several representative real comments from the larger set of survey comments are shown in Fig. 3. The model's CoT reasoning for each step is shown elsewhere, but is omitted here for clarity.

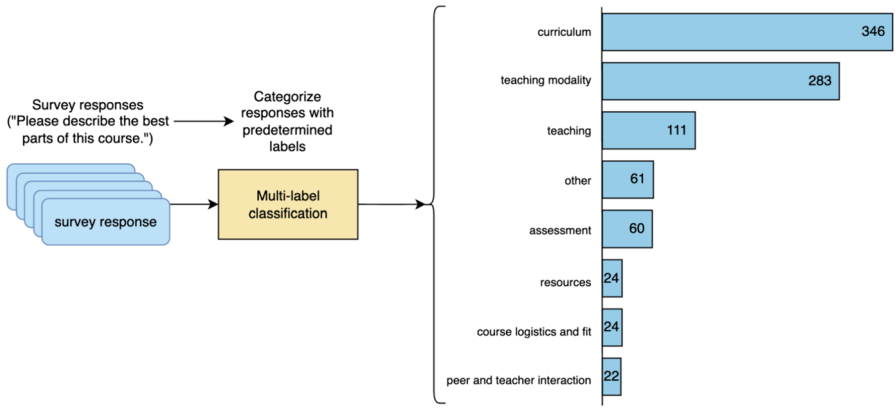

**Fig. 2** Multi-label classification of student comments (results shown using GPT-4)

### Example 4—What Other Content or Topics Were Students Interested in Seeing Covered?

A common goal in analyzing student feedback is to better understand the gaps in course content, in order to decide whether to develop additional material or even new courses. To see if this type of information could reliably be derived from survey responses, we focused on responses to relevant survey questions (Q3 and Q4) for immunology courses with the workflow shown in Fig. 4. Results for several representative real comments are shown. First, just the portions containing new content or topic area suggestions are extracted from the survey responses. Content suggestion themes are then derived and summarized from the excerpts; this is done in batches if they cannot be fit within a single prompt to the LLM (i.e., if there are too many excerpts to fit in the model's maximum context size). Multi-class classification is performed on the excerpts with the themes from each batch. If thematic analysis is done in batches, sets of themes from these batches are then coalesced to arrive at a final set of content themes. The results suggest that GPT-4 is capable of finding content suggestions despite many being specific to the biomedical domain. This may be due to the volume and diversity of the model's pre-training data (although this training mixture has not been disclosed). Immunology is used as an example, but the workflow is not specific to the type of course.

### Example 5—What Feedback Did Students Give About The Teaching and Explanations?

Feedback about teachers and the quality of teaching and explanations in a course is a frequent objective of academic course surveys. Here, we show a workflow where multi-label classification has already been run as an initial step in high-level analysis, and we use the results of that classification as our initial filter to focus on the identified subset of comments related to teaching (corresponding directly to one of the pre-existing labels), with extraction used to further narrow the output of analysis. The workflow, shown in Fig. 5, consisted of multi-label classification, using the pre-existing labels developed (Table 3) followed by extraction of relevant excerpts from the comments that were classified into the 'teaching' category (9% of total comments). If multi-label classification hadn't previously been run, extraction could have been performed on the broader group of comments as the initial step. For our dataset, which includes numerous multi-topic comments, the extraction step was used to further filter the information to only content related to the goal. Results for several representative real comments (de-identified in pre-processing) from the larger set of survey comments are shown in Fig. 5, including one where the model improperly filtered out the comment despite it containing a reference to the quality of explanations. An error such as the one shown could be considered somewhat subtle and highlights the need with zero-shot prompting of LLMs for clear specification of the goal of the extraction.

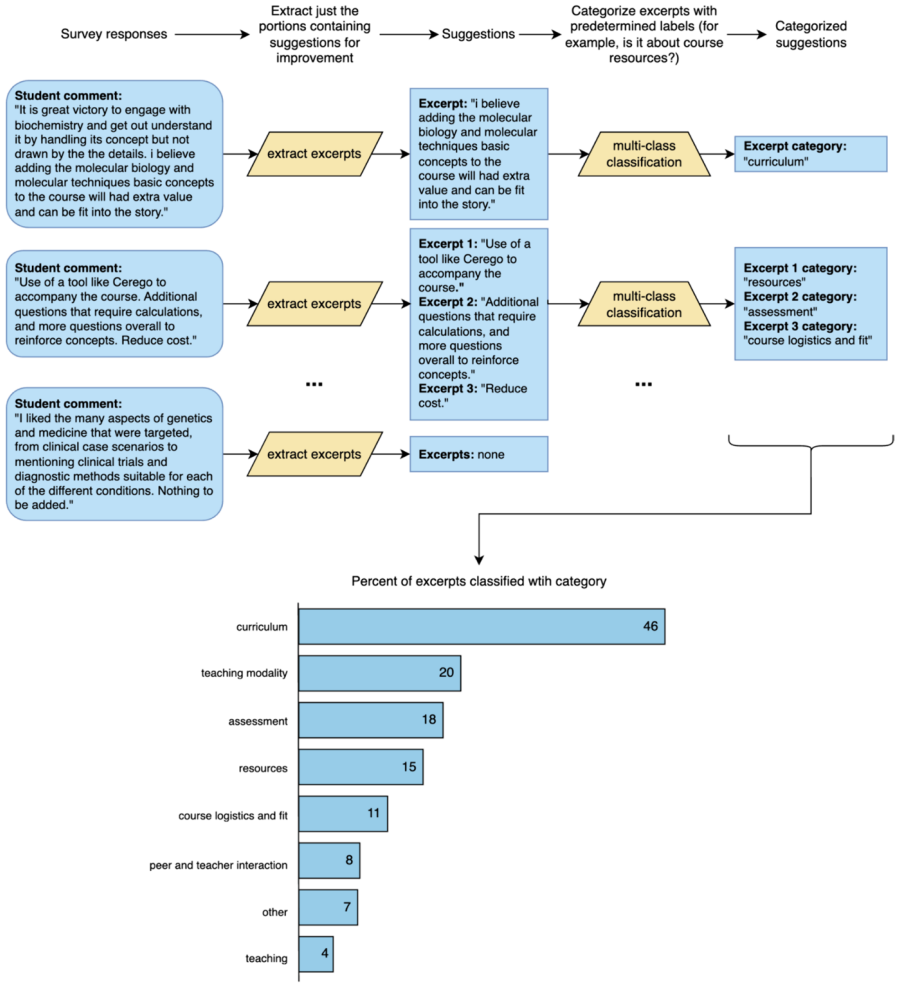

**Fig. 3** Finding suggestions for improvement from student comments (results shown using GPT-4)

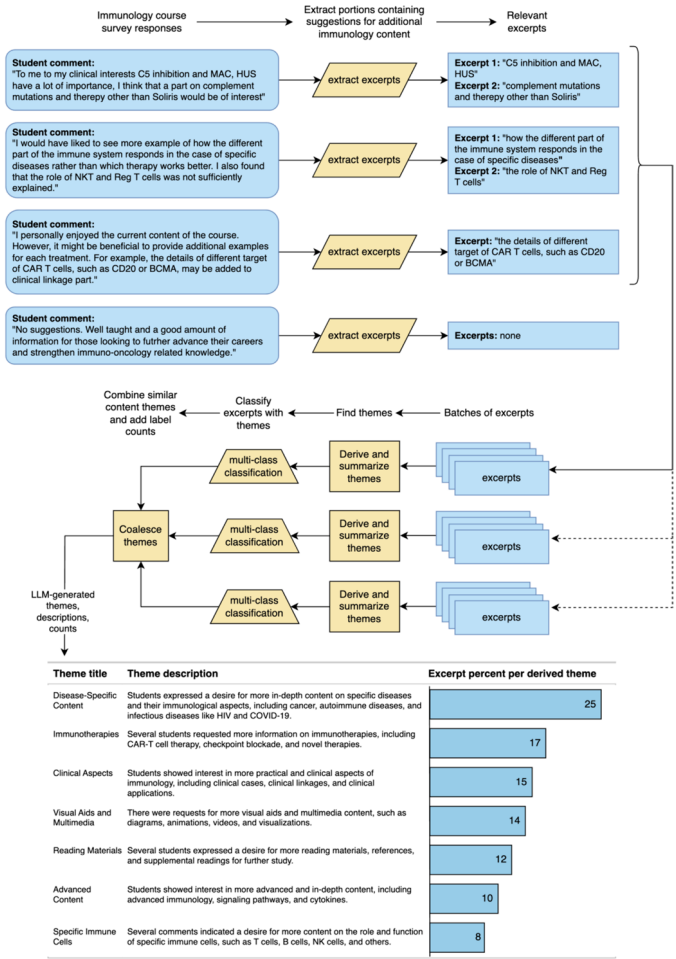

**Fig. 4** Finding suggestions for new immunology content from student comments (results shown using GPT-4)

### Chain-of-Thought Reasoning

The prompts for binary classification, multi-label classification, multi-class classification, sentiment analysis, and evaluation of extraction results all used zero-shot chain-of-thought (CoT) to enhance the quality of the results while maintaining the zero-shot conditions of this study. The CoT reasoning was included in the structured output, allowing for inspection. Only the reasoning from GPT-4 was consistently reliable, and examples are shown here.

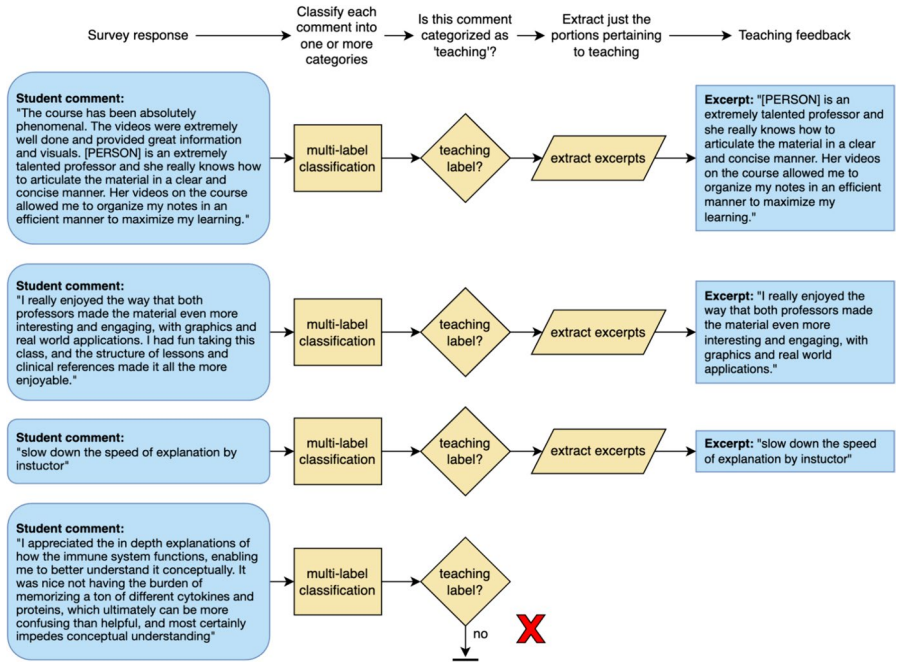

**Fig. 5** Feedback about teaching and explanations (results shown using GPT-4). The red 'x' indicates an error by the model

Example results for binary and multi-class classification tasks are shown in Fig. 6 and Fig. 7, and reasoning for sentiment analysis is also shown in Fig. 7. The reasoning, inspected manually over several hundred comments, is consistent with the classification results and appears to provide logical justification that is grounded in the contextual information (e.g., labels and descriptions) included as part of the prompts (see Electronic Supplement). This suggests that the CoT reasoning from GPT-4 meets a threshold of consistency and logic that allows for potential downstream use cases such as prompt tuning and insight into reasoning for end-users. Potential benefits and caveats of such uses are explored in the Discussion.

Evaluation of the extraction task used a custom LLM evaluation (see Electronic Supplement), developed for this study. In order to refine the evaluation to align results with human preferences, we inspected the CoT reasoning along with the structured eval results for the separate development set of survey responses and made modifications to the evaluation prompts in an iterative fashion. An example of the CoT output for GPT-4 is shown in Fig. 8. As prompts were altered based on human review, the eval results changed in a consistent fashion, suggesting that GPT-4 provided CoT reasoning may be useful in refining LLM evaluations.

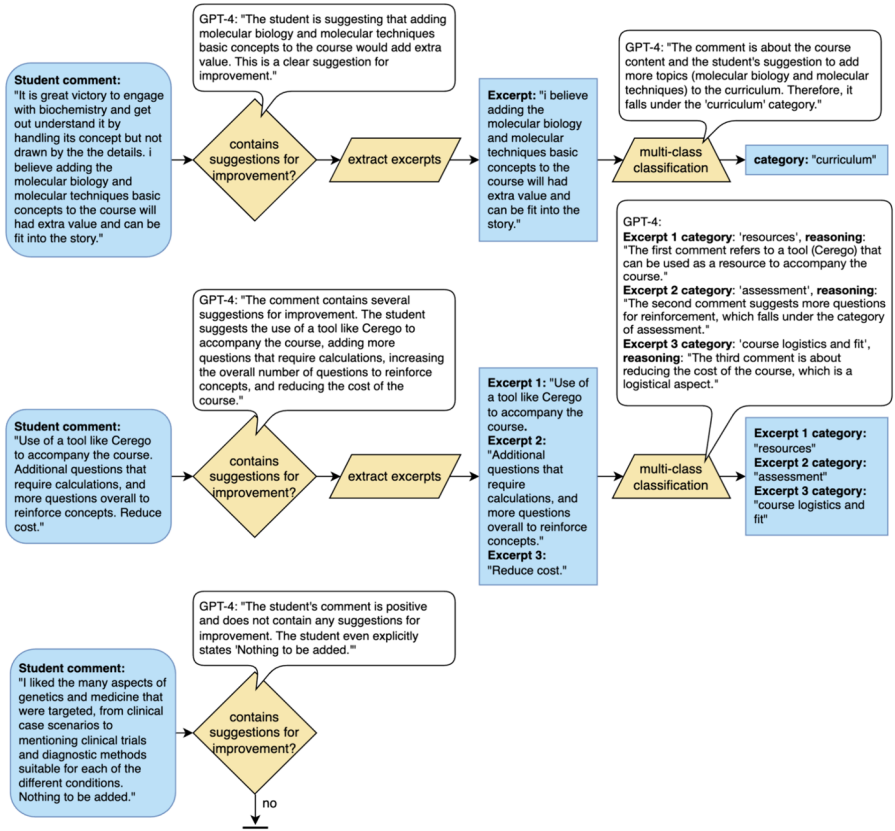

**Fig. 6** Examples of GPT-4 CoT reasoning for binary classification and multi-class classification related to the task of finding suggestions for improvement

## Individual NLP Task Evaluations

To better assess the reliability of workflows such as those shown in the examples and to answer Research Question 3, we evaluated the individual tasks, including multi-label classification, binary classification, extraction, and sentiment analysis.

## Multi-Label Classification Metrics

The difficulty of multi-label classification tasks varies widely (Meidinger & Aßenmacher, 2021), depending on the content to which the labels are being applied, the design of the labels (for example, the clarity of their specification and the potential for overlap), and the number of labels. To put the LLM results in context, we show the inter-rater agreement for application of the eight-label set (Table 3) to our dataset and also compare the LLM results to SetFit, another classification technique.

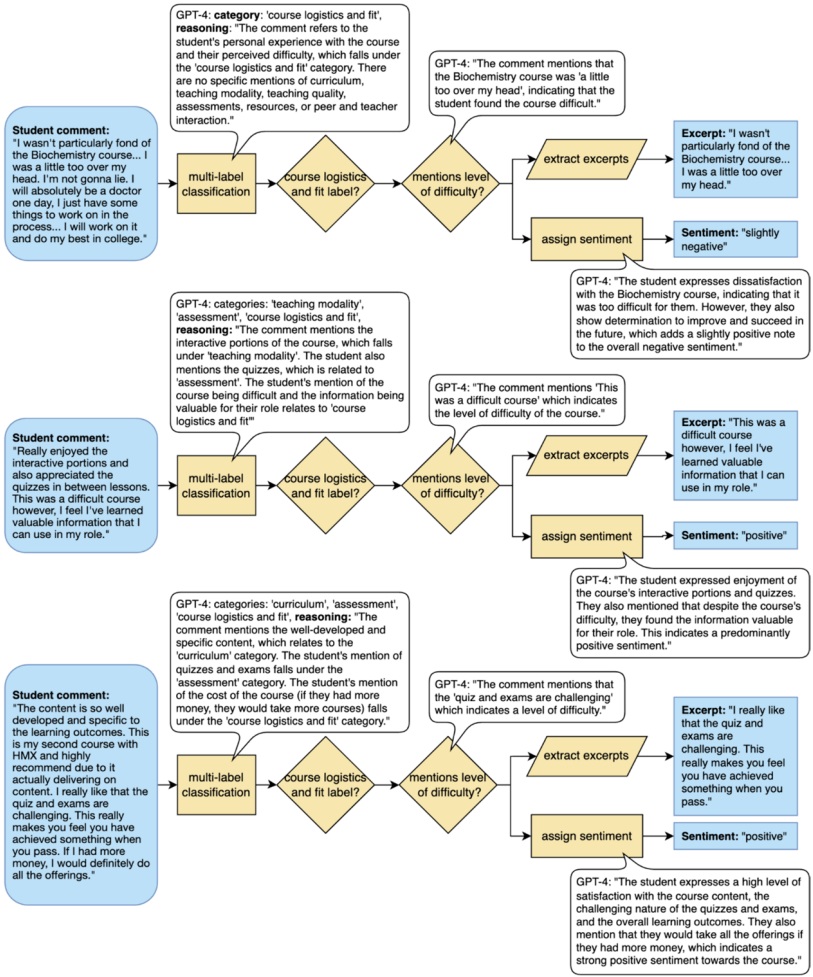

**Fig. 7** Examples of GPT-4 CoT reasoning for multi-label classification, binary classification, and sentiment analysis related to the task of finding how students felt about the level of difficulty of the course

**Inter-rater Agreement** 1,413 (57%) of 2,500 rows had all 4 human raters in agreement across all selected labels and 1,572 (63%) had majority (3 of 4) agreement on all selected labels. The average Jaccard similarity coefficient including all 2,500 rows (averaged across the six unique pairings of the four human raters for all rows) was 81.24% (Table 4), suggesting that this was a challenging task even for expert human annotators who developed the custom label set in close collaboration. GPT-4 agreement with human annotators is shown; the average across all pairings including GPT-4 was 80.60%.

**LLM and SetFit Evaluation** In addition to evaluating the GPT models, we also performed multi-label classification using SetFit (Tables 5 and 6).

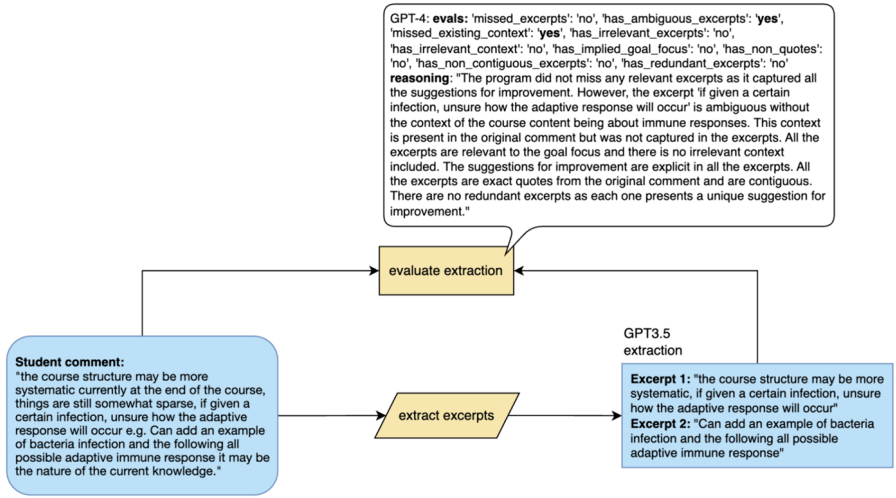

**Fig. 8** Example of GPT-4 CoT reasoning for extraction evaluation

**Table 4** Inter-rater Jaccard similarity coefficients, including human annotators and GPT-4 as another rater/annotator (human pairs average = 81.24%; all pairs average = 80.60%)

|  | Annotator1 | Annotator2 | GPT-4 | Annotator3 | Annotator4 |
| --- | --- | --- | --- | --- | --- |
| Annotator1 | - | 81.27 | 80.18 | 83.37 | 82.35 |
| Annotator2 | 81.27 | - | 79.40 | 80.84 | 78.42 |
| GPT-4 | 80.18 | 79.40 | - | 80.74 | 78.22 |
| Annotator3 | 83.37 | 80.84 | 80.74 | - | 81.18 |
| Annotator4 | 82.35 | 78.42 | 78.22 | 81.18 | - |

**Table 5** Evaluation on consensus rows, with majority agreement on all tags (1,572 rows for LLMs, 1,489 rows for SetFit)

|  | Jaccard | Average precision | Macro average | | | Micro average | | |
| --- | --- | --- | --- | --- | --- | --- | --- | --- |
|  |  |  | Precision | Recall | F1 | Precision | Recall | F1 |
| GPT-4 | 92.97 | 93.91 | 89.88 | 90.59 | 89.78 | 93.66 | 93.26 | 93.46 |
| GPT-3.5 | 72.61 | 74.79 | 69.34 | 82.18 | 72.63 | 72.36 | 84.48 | 77.96 |
| SetFit | 73.86 | 78.01 | 84.37 | 57.59 | 66.85 | 91.92 | 71.43 | 80.39 |

For the consensus rows evaluation, the zero-shot results for GPT-4 are similar to what might be expected of fine-tuned classifiers (Meidinger & Aßenmacher, 2021). The other models have strengths and weaknesses, with SetFit having relatively high precision and lower recall, and GPT-3.5 following the converse pattern. The overall results for SetFit and GPT-3.5, focusing on Jaccard coefficient and F1 scores, are similar. The results emphasize 1) the fact that fine-tuning is desirable when feasible, approaching the performance of powerful LLMs like GPT-3.5 even with a few-shot fine-tuning approach; and 2) the quality of the zero-shot performance of GPT-4.

**Table 6** Evaluation on all rows using consensus labels (2500 rows for LLMs, 2359 rows for SetFit)

|  | Jaccard | Average precision | Macro average | | | Micro average | | |
| --- | --- | --- | --- | --- | --- | --- | --- | --- |
|  |  |  | Precision | Recall | F1 | Precision | Recall | F1 |
| GPT-4 | 80.17 | 81.53 | 73.91 | 88.38 | 79.69 | 78.32 | 89.70 | 83.63 |
| GPT-3.5 | 63.00 | 65.18 | 60.42 | 79.79 | 65.75 | 60.31 | 83.45 | 70.02 |
| SetFit | 62.72 | 67.52 | 73.22 | 53.08 | 59.61 | 79.40 | 65.14 | 71.57 |

**Binary Classification Metrics**

1,250 comments (randomly taken from the original selected 2,500 comments, evenly distributed across Q1 through Q4) were classified based on the question "does this comment contain suggestions for improvement?", with the model assigning a 'yes' or 'no' to each comment, and results were compared against one expert human annotator—as the gold standard—assigning 'yes' or 'no' to each comment. Binary classification could be considered the simplest of the evaluated NLP tasks, and both LLM models exhibited good performance (Table 7).

**Extraction Evaluation**

Using 'suggestions for improvement' as an example target of extraction, comments were first classified via GPT-4 as containing the target or not (see binary classification task above). Of the 1,250 comments, 716 were labeled by binary classification as containing suggestions for improvement. These comments were then run through extraction to find the individual excerpts. The quality of the extraction was then scored by the following method, employing the concept of LLM-as-a-judge (Huang et al., 2024) where custom LLM-based evaluations are used for cases where standardized evaluation methods don't exist or are inadequate. For each comment's excerpt(s), GPT-4 was used to apply a custom evaluation rubric with nine questions. Only GPT-4 was capable of applying the evaluation reliably and was therefore used. An example of one of the rubric questions was "Did the program extract any irrelevant excerpts? (yes or no)" (see the extraction evaluation prompt in the Electronic Supplement for all questions in the rubric). That question is abbreviated as the "Irrelevant Excerpts" category in the table below, showing the percentage of "yes" answers. Given that each of the rubric questions had a yes/no answer, with a "yes" answer indicating a failure of the aspect of the model's extraction based on that question, the error rate or percentage of failures for each aspect could be determined. The extracted excerpts were also examined by a human annotator to determine the percentage of the 716 rows that contained hallucinations in the excerpts, as defined by substantial edits or complete fabrication of additional language not present in the original comment. That percentage is shown in the last row of Table 8, with the other rows' results representing the scores (error rates) determined by GPT-4 judging the extraction results from GPT-4 and GPT-3.5.

**Table 7** Binary classification task performance

| Model | Accuracy | Precision | Recall | F1 |
| --- | --- | --- | --- | --- |
| GPT-4 | 95.20 | 96.20 | 95.39 | 95.79 |
| GPT-3.5 | 90.16 | 89.01 | 93.35 | 91.14 |

**Table 8** Error rate (%) of extraction for 'suggestions for improvement' from comments classified as containing 'suggestions for improvement' (all metrics from rubric and human annotation for hallucinations)

| Metric | GPT-4 | GPT-3.5 |
|---|---|---|
| Missed Excerpts | 2.37 | 7.82 |
| Ambiguous Excerpts | 4.61 | 4.75 |
| Missed Existing Context | 0.28 | 0.84 |
| Irrelevant Excerpts | 0.14 | 0.84 |
| Irrelevant Context | 0.00 | 0.14 |
| Implied Goal Focus | 3.07 | 2.79 |
| Non-Quotes | 0.00 | 6.01 |
| Non-Contiguous Excerpts | 0.00 | 0.14 |
| Redundant Excerpts | 0.28 | 2.79 |
| Hallucinations | 0.00 | 3.91 |

**Table 9** Classification of comments as negative, positive, or neutral relative to human annotator

| Model | Accuracy | Precision (macro) | Recall (macro) | F1 (macro) |
|---|---|---|---|---|
| GPT-4 | 80.86 | 82.65 | 80.28 | 80.78 |
| GPT-3.5 | 65.17 | 73.68 | 66.44 | 64.88 |
| twitter-roberta-base-sentiment-latest | 66.69 | 71.38 | 64.86 | 61.10 |

The GPT-4 model included some ambiguous excerpts; however, those were most commonly due to lack of context in the comment itself, rather than the model failing to extract that context. GPT-4 followed directions very closely, and its results did not contain hallucinations. In contrast, the output of GPT-3.5 contained hallucinations at a rate of about 4% and edits to comments at a rate of about 6%. GPT-3.5 also missed relevant excerpts significantly more frequently than GPT-4. Additional prompt tuning may reduce the rate of these errors; nonetheless, the results suggest that a degree of caution should be applied in using GPT-3.5 for extraction.

**Sentiment Analysis Metrics**

Using GPT-4 and GPT-3.5, comments related to course suggestions and improvement (Q3 and Q4) were classified as 'negative', 'slightly negative', 'neutral', 'slightly positive', or 'positive'. Table 9 shows accuracy, and macro precision, recall, and F1 scores for three models; comparison was made by grouping 'negative' and 'slightly negative' into a single negative class, keeping 'neutral' as its own class, and grouping 'positive' and 'slightly positive' into a single positive class.

GPT-4 is substantially better on each metric than the other models; however, the results are lower than what has been seen for fine-tuned models on in-domain datasets, indicating that the sentiment expressed in student course feedback may differ from the range of sentiment expressed in the internet training data of these models. The negative class was the most challenging for all models, suggesting that negative course feedback may differ significantly from negative internet feedback.

**LLM Cost and Time**

The cost of using the OpenAI APIs for GPT-4 and GPT-3.5 depends on the number of prompt tokens and number of completion tokens. For the final prompts and tasks used in this study, the average price of running 100 comments is shown in Table 10 for each model for different tasks (cost as of June 2023, when the results were run). The tasks include the CoT reasoning in the completion (output), significantly increasing the number of completion tokens. These provide an approximate gauge given that comments vary in length. Total API cost for this study including prompt tuning was approximately $300.

The time for model calls for GPT-4, the slower of the OpenAI models, was approximately 10 s for running 100 comments in parallel for most tasks listed. For the extraction evaluation, it took approximately 1 min to run 100 comments in parallel. For batches, sleep intervals were also incorporated to stay conservatively within maximum token rates. A small percentage of API calls received errors and automatic retries were used after wait intervals.

## Discussion

Analysis of education feedback, in the form of unstructured data from survey responses, is a staple for improvement of courses (Diaz et al., 2022; Flodén, 2017; Marsh & Roche, 1993; Moss & Hendry, 2002; Schulz et al., 2014). However, this task can be time-consuming, costly, and imprecise, hampering the ability for educators and other stakeholders to make decisions based on insights from the data (Shaik et al., 2023; Mattimoe et al., 2021; Nanda et al., 2021; Shah & Pabel, 2019). Large language models with generative AI capabilities have become widely accessible recently but remain underexplored in analysis of qualitative data from educational feedback surveys.

**Table 10** Cost per 100 comments for GPT-4 and GPT-3.5

| Task | GPT-4 | GPT-3.5 |
| --- | --- | --- |
| Binary classification | $0.93 | $0.04 |
| Multi-label classification | $2.63 | $0.12 |
| Multi-class classification | $2.13 | $0.10 |
| Text extraction | $1.10 | $0.05 |
| Text extraction evaluation | $3.01 | $0.13 |
| Sentiment analysis | $1.17 | $0.05 |
| Inductive thematic analysis | $0.13 | $0.006 |

Our research questions focused on whether these new tools can 1) be successfully used perform a wide variety of natural language processing tasks on survey results (RQ1); 2) offer a degree of transparency based on capturing chain-of-thought intermediate output (RQ2); and 3) perform at a level comparable to human performance across all tasks without needing to be provided with hand-labeled examples (RQ3).

We were able to create reliable, reproducible workflows that put together multiple analysis tasks, running them in minutes on more than one thousand survey responses. The results demonstrate that these workflows can provide insight into a variety of questions that may be asked by educators, including finding suggestions for improvement, identifying themes in students' feedback, and quantifying such results through classification, including multi-label classification. The zero-shot approach (no hand-labeled examples, other than for evaluation metrics) provides flexibility; in other words, tasks can rapidly be adjusted through changes in the LLM prompts, and new tasks (for example, using extraction to find information about a different focus) can be added without need for model fine-tuning or labeling new examples.

We show that chain-of-thought prompting, which was used to increase accuracy, may also provide insight into the model's reasoning or trajectory. It is possible that the LLM is imitating plausible reasoning rather than providing insight into how it actually arrived at its answer; however, this distinction may be immaterial given that 1) GPT-4's reasoning was logical and highly consistent with the results, displaying elements of causal reasoning; and 2) when prompts were changed, reasoning results changed accordingly. This has been discussed in recent work; GPT-4 has been shown to score highly on evaluations of causal reasoning (Kıcıman et al., 2023). In Peng et al., 2023, GPT-4 was used for evaluation of other LLMs and was able to provide consistent scores as well as detailed explanations for those scores. While specialized non-LLM models can provide signals like confidence scores in individual classes, they lack more detailed explanations of results; we believe that seeing a version of logical reasoning behind complex output can foster confidence and reduce the perception of these models as black boxes. Furthermore, it is important to consider that having human annotators reliably provide consistent, logical justification for each annotation is prohibitive for datasets of any appreciable scale.

The evaluation metrics for each individual task show a human or near-human level of performance across a range of tasks for GPT-4. GPT-3.5 does not reach this level of accuracy. Our tasks and dataset were drawn from real-world data and actual use cases, with some of the tasks (e.g., multi-label classification) proving challenging even for expert annotators. The human-like level of GPT-4 can be seen in examples of the reasoning results as well (Figs. 6, 7, and 8). In addition, it outperformed label-efficient fine-tuned classifiers like SetFit. For workflows that chain together two or more NLP tasks, like those examined in this study, it is important that the performance on each task is reliable enough such that errors do not accrue in the process of obtaining a final result. It is likely that at the time of publication there are other models (e.g., the latest version of Claude or the top-performing open-source models) that perform at similar or higher levels than the version of GPT-4 used in this study.

Overall, we found that large language models are at a stage where they can be effectively used across the broad range of tasks that are part of survey analysis. Use of LLMs for this purpose has implications for education. These implications include:

1) the potential to democratize access to high quality qualitative survey analysis. The paradigm of "language in, language out" allows non-machine learning experts to create workflows that can handle a range of tasks.
2) a drastic reduction in the time and effort involved for challenging survey analysis tasks, shortening the feedback loop. In many cases where there is a high volume of survey responses, the effort and expertise necessary to arrive at data-driven results may otherwise be prohibitive. The goal is to be able to analyze unstructured text survey data as easily as one might analyze quantitative results like those from Likert scales.
3) making educators and education leaders more ambitious in terms of the questions they can answer based on student feedback. For example, doing thematic analysis on a volume of survey responses may have been infeasible previously.

We look forward to people using these new tools to effectively learn from student feedback.

## Limitations

The data used in this study was from a specific domain (online biomedical science courses) and was in English. Other domains and languages were not tested.

LLM results were highly prompt-dependent, and others may achieve even more accurate results than those we have shown. Even within the most capable models, we observed that prompting techniques and prompt tuning made a significant difference. There is considerable literature on effective methods of prompting. There is an interplay of prompting techniques with the behavior of instruction-tuned models in a way that may or may not fully elicit the capabilities of each model, with prompts being seen as a form of hyperparameter to the model and with responses changing depending on updates to model training (Chen et al., 2023).

Other than the comparison to SetFit and to the RoBERTa sentiment analysis model, we limited our exploration to recent OpenAI models; future work may expand this to include other models such as the most capable proprietary models (e.g., Claude and Gemini), and the most capable open-source models.

## Concluding Remarks

While LLM analysis approaches are being used in other fields, like customer reviews and user feedback (Morbidoni, 2023; Abdali et al., 2023), there has not yet been rigorous demonstration of their utility and accuracy in student feedback surveys. Our contributions are to demonstrate the viability of using LLMs for this purpose, and to perform a thorough analysis of the feasibility and evaluation of the quality of LLMs' results to show reliability in common qualitative survey analysis tasks.

Future research could incorporate additional prompting techniques to improve capabilities and accuracy. For example, self-consistency (Wang et al., 2022), reflection (iterative self-refinement, Madaan et al., 2023), and few-shot learning have been studied and shown to be reliable means of improving performance on difficult tasks; these were out-of-scope for the zero-shot premise of this article but are worth exploring. In addition, the ability to compose survey analysis workflows is also amenable to the use of agents (Shen et al., 2023; Weng, 2023; Yao et al., 2022). An educator or other stakeholder analyzing survey feedback should be able to state a goal or intent to an LLM-powered agent, with the agent picking and running tasks as a chain to get the desired analysis. Such an agent could also incorporate non-LLM tools, for example if a fine-tuned model is available that excels on a given task and is well-matched to the dataset at hand. Ideally, users of such tools should be able to operate by stating intent rather than tuning prompts or fine-tuning specialized models.

We look forward to future progress in exploring the capabilities of these new tools; the models will continue to improve but even in their current state, they can be powerful tools for survey analysis.

# Appendix

## Additional Metrics for Multi-Label Classification

### Consensus Rows—1572 Rows Dataset (1489 for SetFit)

Precision, recall, and F1 score are shown for each tag in multi-label classification for the consensus rows condition, along with macro averages for each metric, in Table 11 for GPT-4. Hamming loss and subset accuracy are shown in Table 12.

**Table 11** Individual label scores for multi-label classification with GPT-4, consensus rows

| Tag | Precision | Recall | F1 Score |
| --- | --- | --- | --- |
| Course logistics and fit | 89.83 | 64.63 | 75.18 |
| Curriculum | 95.89 | 91.48 | 93.64 |
| Teaching modality | 97.78 | 96.70 | 97.24 |
| Teaching | 76.04 | 91.25 | 82.95 |
| Assessment | 97.50 | 93.41 | 95.41 |
| Resources | 92.31 | 97.30 | 94.74 |
| Peer and teacher interaction | 77.08 | 92.50 | 84.09 |
| Other | 92.60 | 97.47 | 94.97 |
| Macro average | 89.88 | 90.59 | 89.78 |

**Table 12** Hamming loss and subset accuracy for multi-label classification, consensus rows

| Model | Hamming loss | Subset accuracy |
| --- | --- | --- |
| GPT-4 | 0.0194 | 0.8849 |
| GPT-3.5 | 0.0710 | 0.5948 |
| SetFit | 0.0508 | 0.6797 |

**Table 13** Individual label scores for multi-label classification with GPT-4, consensus labels

| Tag | Precision | Recall | F1 Score |
| --- | --- | --- | --- |
| Course logistics and fit | 74.81 | 57.71 | 65.16 |
| Curriculum | 82.86 | 86.38 | 84.58 |
| Teaching modality | 78.35 | 95.94 | 86.26 |
| Teaching | 56.70 | 87.59 | 68.83 |
| Assessment | 87.36 | 94.82 | 90.94 |
| Resources | 70.95 | 95.49 | 81.41 |
| Peer and teacher interaction | 60.58 | 94.03 | 73.68 |
| Other | 79.64 | 95.10 | 86.68 |
| Macro average | 73.91 | 88.38 | 79.69 |

**Table 14** Hamming loss and subset accuracy for multi-label classification, consensus labels

| Model | Hamming loss | Subset accuracy |
| --- | --- | --- |
| GPT-4 | 0.0503 | 0.7168 |
| GPT-3.5 | 0.10235 | 0.4628 |
| SetFit | 0.07238 | 0.5774 |

**Consensus Labels—2500 Rows Dataset (2359 for SetFit)**

Precision, recall, and F1 score are shown for each tag in multi-label classification for the consensus labels condition, along with macro averages for each metric, in Table 13 for GPT-4. Hamming loss and subset accuracy are shown in Table 14.

**Supplementary Information** The online version contains supplementary material available at https://doi.org/10.1007/s40593-024-00414-0.

**Acknowledgements** We wish to thank members of the HMX team for their contributions to creating and administering the surveys used in this study.

**Author Contributions** Conceptualization, methodology, software, analysis, and drafting of the manuscript were performed by Michael J. Parker. Development of labels, annotation/labeling for multi-class classification, and refinement of the manuscript were performed by all authors (equal contributions).

**Funding** This research received no external funding.

**Availability of Data and Materials** The prompts and the function schemas used in this study are shared in supplementary material (Electronic Supplements 1 and 2). To help preserve the anonymity of students and of feedback about teachers, the survey responses are not included in an open-access repository. The data may be provided upon request to the authors and approval of the university research ethics board.

## Declarations

**Ethics Approval and Consent to Participate** This study was determined not to be human subjects research by the Harvard Medical School Office of Human Research Administration.

**Competing Interests** The authors have no relevant financial or non-financial interests to disclose.

**Open Access** This article is licensed under a Creative Commons Attribution 4.0 International License, which permits use, sharing, adaptation, distribution and reproduction in any medium or format, as long as you give appropriate credit to the original author(s) and the source, provide a link to the Creative Commons licence, and indicate if changes were made. The images or other third party material in this article are included in the article's Creative Commons licence, unless indicated otherwise in a credit line to the material. If material is not included in the article's Creative Commons licence and your intended use is not permitted by statutory regulation or exceeds the permitted use, you will need to obtain permission directly from the copyright holder. To view a copy of this licence, visit http://creativecommons.org/licenses/by/4.0/.

## Authors and Affiliations

### Michael J. Parker[1]  · Caitlin Anderson[1] · Claire Stone[2] · YeaRim Oh[2]

✉ Michael J. Parker
michael_parker@hms.harvard.edu

Caitlin Anderson
caitlin_anderson@hms.harvard.edu

Claire Stone
claire_stone@hms.harvard.edu

YeaRim Oh
yearim_oh@hms.harvard.edu

[1] Office of Online Learning, Harvard Medical School, HMX, 4 Blackfan Circle, Boston, MA, USA

[2] Office of External Education, Harvard Medical School, Boston, MA, USA